\documentclass{article} %
\usepackage{iclr2027_conference,times}

\usepackage{amsmath,amsfonts,bm}

\def\eqref#1{equation~\ref{#1}}

\def\1{\bm{1}}

\DeclareMathAlphabet{\mathsfit}{\encodingdefault}{\sfdefault}{m}{sl}
\SetMathAlphabet{\mathsfit}{bold}{\encodingdefault}{\sfdefault}{bx}{n}

\usepackage{amssymb}
\ifdefined\pdfminorversion
\fi

\usepackage{xcolor}
\usepackage{hyperref}
\definecolor{cornellred}{rgb}{0.7, 0.11, 0.11}
\definecolor{cadmiumgreen}{rgb}{0.0, 0.42, 0.24}
\definecolor{Blue9}{rgb}{0.098,0.3,0.9}
\definecolor{projpink}{RGB}{231,84,128}
\newcommand{\projecturl}{https://senri-video.github.io/}
\hypersetup{
  linkcolor = cornellred,
  citecolor  = cadmiumgreen,
  colorlinks = true,
  urlcolor = Blue9
}
\usepackage{url}
\usepackage{graphicx}
\usepackage{booktabs}
\usepackage{multirow}
\usepackage{makecell}
\usepackage{pifont}
\usepackage{capt-of}
\usepackage{enumitem}
\usepackage{placeins}
\usepackage{longtable}
\usepackage{array}
\graphicspath{{figs/}}

\newcommand{\cmark}{\ding{51}}
\newcommand{\xmark}{\ding{55}}
\newcommand{\ck}{\textcolor{cadmiumgreen}{\cmark}}
\newcommand{\xk}{\textcolor{cornellred}{\xmark}}

\title{SENRI: Understanding the Space--Event Nexus in Real-Life Videos}

\iclrfinalcopy

\author{Kehan Lan$^{1,2}$\quad Kaining Ying$^{1}$\quad Henghui Ding$^{1}$\quad Kaipeng Zhang$^{2}$\thanks{Corresponding author.} \\
$^{1}$Fudan University\quad $^{2}$Alaya Lab
}

\begin{document}

\maketitle

\begin{abstract}
Human memories are organized in a way that connects events with their corresponding spatial contexts. An AI assistant should also be able to memorize and recall the nexus of events and spaces. Thus we introduce \textbf{SENRI}, a task that studies this space--event connection in long, continuous real-life egocentric video. It contains six task types in two directions: \textit{Event$\rightarrow$Space}, which recovers spatial knowledge from event cues, and \textit{Space$\rightarrow$Event}, which retrieves events from regions and trajectories. To evaluate this, we build \textbf{SENRI-Bench}, containing 500 \textit{manually annotated} multiple-choice questions over 10 uninterrupted \textit{self-recorded} videos with around 25.3 hours of content. Also, we evaluate three typical long video understanding paradigms: direct perception, evidence seeking, and persistent memory. The task remains challenging across all three paradigms, highlighting substantial room for improvement. Then we show two representative failure modes of existing memory methods and address these evidence gaps with \textbf{SENRI-Agent}, a controller that uses \textit{Target-Guided Visual Check (TGVC)} to propose local visual check tasks for parallel verifiers and integrate the reports into a working memory to help with reasoning. Built upon existing memory systems, SENRI-Agent improves the performance of existing baselines without modifying their stored memories. We hope SENRI will provide valuable insights into how real-life egocentric understanding should be evaluated, reveal the gaps in existing approaches, and offer a possible direction for improvement.  \href{\projecturl}{\textcolor{projpink}{\textbf{Project Page.}}}
\end{abstract}

\section{Introduction}

The world we inhabit is a continuous weaving of the spaces and events. As we move through it from an egocentric perspective, what happens becomes inseparable from how we experience the space where it happens. Our memories preserve the combination of the two. We can effortlessly ask, “Where was I when I tasted the fruit flower tea?” or “What did I encounter after walking to the end of the canal and returning along the opposite side?” and expect answers rich enough to reconstruct the experience in the mind’s eye. This ability to bind what happened with where it happened is fundamental to how we remember and understand lived experience. We therefore ask: can an AI assistant reason over the same \emph{space--event nexus} from real-life videos?

\begin{table}[!t]
    \centering
    \setlength{\abovecaptionskip}{0pt}
    \caption{
        Comparison of SENRI-Bench with existing video understanding benchmarks.
        \ck{} indicates coverage or partial coverage;
        \xk{} indicates absence or no explicit guarantee.
        \textit{Anno.} denotes the QA annotation procedure: A for automatic,
        M for manual, and A/M for a mixed pipeline.
        \textit{Self-recorded} denotes videos newly collected for the project
        and absent from existing resources.
        \textit{Real-life} denotes everyday activities or exploration outside
        dedicated indoor scans.
        \textit{Event-grounded} includes questions about events and \textit{Spatial-reasoning} requires reasoning about spatial conditions.
    }
    \label{tab:benchmark_comparison}
    \small
    \setlength{\tabcolsep}{4pt}
    \renewcommand{\arraystretch}{0.95}

    \resizebox{\textwidth}{!}{%
    \begin{tabular}{@{}llcccccccr@{}}
        \toprule
        Category
        & Benchmark
        & Anno.
        & \makecell{Self-\\recorded}
        & \makecell{One-\\take}
        & \makecell{Real-\\life}
        & \makecell{Cross-\\scene}
        & \makecell{Event-\\grounded}
        & \makecell{Spatial\\reasoning}
        & \makecell{Avg. duration\\per clip (h)} \\
        \midrule

        \multirow{4}{*}{\makecell[l]{Spatial\\understanding\\benchmarks}}
        & VSI-Bench
        & A/M
        & \xk & \xk & \xk & \xk & \xk & \ck & $\approx 0.03$ \\

        & VSI-Super
        & A/M
        & \xk & \xk & \xk & \ck & \xk & \ck & $0.17$--$4.00$ \\

        & VSI-Super-Wild
        & A/M
        & \xk & \ck & \ck & \ck & \xk & \ck & $0.64$ \\

        & UCS-Bench
        & A/M
        & \xk & \xk & \ck & \ck & \ck & \ck & $0.33$ \\

        \midrule

        \multirow{5}{*}{\makecell[l]{Long video\\benchmarks}}
        & LVBench
        & M
        & \xk & \xk & \ck & \ck & \ck & \xk & $1.14$ \\

        & LongVideoBench
        & M
        & \xk & \xk & \ck & \ck & \ck & \xk & $0.13$ \\

        & HippoVlog
        & A/M
        & \xk & \xk & \ck & \ck & \ck & \xk & $0.46$ \\

        & HourVideo
        & A/M
        & \xk & \ck & \ck & \ck & \ck & \ck & $0.76$ \\

        & EgoLifeQA
        & A/M
        & \ck & \ck & \ck & \ck & \ck & \xk & $44.3$ \\

        \midrule

        Ours
        & \textbf{SENRI-Bench}
        & \textbf{M}
        & \ck
        & \ck
        & \ck
        & \ck
        & \ck
        & \ck
        & \textbf{2.53} \\

        \bottomrule
    \end{tabular}%
    }
\vspace{-3mm}

\end{table}

\begin{figure*}[!t]
    \centering
    \input{figs/task_distribution}

    \vspace{3mm}
    
    \input{figs/benchmark_examples}
\end{figure*}

None of existing benchmarks fully evaluate this. As
Table~\ref{tab:benchmark_comparison} shows, spatial-understanding benchmarks test spatial tasks such as relations and counting, yet they seldom ground spatial questions in events or consist exclusively of uncut, one-take videos captured in real-life environments~\citep{vsibench,cambrians,vsisuperwild,ucsbench}. Long-video benchmarks provide real-life footage and emphasize event understanding and long range recall. But they neither consistently evaluate spatial reasoning~\citep{lvbench,longvideobench,hippomm,egolife} nor establish the evaluation of the connection between space and events in newly recorded videos ~\citep{hourvideo}. This gap leaves the nexus between what happened and where it happened largely untested and motivates us to work on a new perspective of long video understanding.

To this end, we introduce \textbf{SENRI}, a task for understanding the \textbf{S}pace--\textbf{E}vent \textbf{N}exus in \textbf{R}eal-L\textbf{I}fe videos. SENRI has two complementary directions. \textit{Event$\rightarrow$Space (ES)} starts from an experienced or observed event and asks about its spatial context through \emph{localization, relation reasoning, object counting,} and \emph{navigation planning}. \textit{Space$\rightarrow$Event (SE)} starts from a region or route and asks what happened there through \emph{region-conditioned event retrieval (RER)} and \emph{trajectory event retrieval (TER)}. The six task types require models to establish a deep understanding of the real world in an egocentric perspective and reason about the relationship between events and spatial contexts. Then we build SENRI-Bench from 10 \emph{self-recorded} videos totaling 25.3 hours and 500 \emph{manually annotated} multiple-choice questions. Each video is a single uninterrupted take, preserving movement between scenes and the temporal order of events. The person who records a video also writes its questions. First-hand knowledge of the environment and events helps the annotator design more valuable questions. Each video contributes 50 questions and serves as a single evaluation episode.

Then we evaluate 28 configurations across three long video understanding paradigms for long video understanding: direct perception that directly consumes sampled frames~\citep{videochat3,internvideo3,cambrians,qwen3.5,qwen36_27b,qwen36_35b_a3b,qwen38,kimik3,gemini31pro,gemini35flash,gemini37flash,openai2025gpt5,openai2026gpt55}, evidence seeking agents that reason over a long video by integrating tools~\citep{videoo3,longvideor1,paravt,videoarm}, and persistent memory systems that first build memories then perform querying and reasoning over memories~\citep{m3agent,lightomni,worldmm,egagent,magicvideo}. It turns out that the strongest direct perception, evidence seeking, and persistent memory configurations only reach 52.2\%, 59.0\%, and 58.8\% accuracy respectively. Moreover, our findings include that methods' performance differs across different tasks and the inference model also has a large effect on persistent memory systems. Those findings suggest that the space--event nexus is a challenging task for existing long video understanding methods and implies the possible gaps in current systems.

Also, we show two representative failure modes in existing memory methods in a case study: weak space–event nexus in text memory and missing target details in visual memory. That motivates us to develop SENRI-Agent, a controller that builds upon previous search and answer workflow but further introduces \textit{Target-Guided Visual Check (TGVC)} which first specifies the event and spatial context implied by retrieved memories, then assigns targeted visual checks to parallel visual verifiers. Verifiers return targeted finding reports which are further used to update the working memory, allowing the reasoner to perform more targeted verification. With the inference models and stored memories held fixed, SENRI-Agent improves from two baselines' performance, without rebuilding their memories or modifying retrieving interfaces. In a nutshell, our contributions are threefold:

\begin{itemize}[leftmargin=*,labelsep=0.01em,topsep=2pt]
    \item We propose SENRI to formulate understanding space--event nexus as a new long-video understanding
    problem with two complementary directions: Event$\rightarrow$Space and Space$\rightarrow$Event, and six different task types: localization, navigation planning, object counting, region-conditioned event retrieval, relation reasoning, and trajectory event retrieval.

    \item We instantiate
    it in SENRI-Bench, which contains 500 manually annotated questions over
    25.3 hours of self-recorded, uninterrupted egocentric video, evaluate three long video understanding paradigms on SENRI-Bench, and reveal the gaps in existing methods.

    \item We further propose SENRI-Agent, which uses TGVC and working memory to improve two persistent memory systems without modifying their memories or retrieval interfaces.
\end{itemize}

\section{SENRI-Bench}

\label{sec:benchmark}

\paragraph{Task Formulation and Taxonomy.} 
Each SENRI-Bench example pairs a long, continuous egocentric video with a four-choice question that connects an event to its spatial context. The questions follow two reciprocal directions. Event$\rightarrow$Space (ES) uses an event as the retrieval cue and queries its spatial context through four task types: \emph{localization} identifies where an event occurred; \emph{relation reasoning} recovers spatial relations among entities or places; \emph{object counting} counts distinct entities within the context; and \emph{navigation planning} recovers or plans a route. Space$\rightarrow$Event (SE) uses spatial context as the cue and asks the model to retrieve what occurred there: \emph{region-conditioned event retrieval (RER)} queries events associated with a specified place or region; \emph{trajectory event retrieval} queries events encountered along a described route or trajectory.

 \paragraph{Data Collection, Annotation, and Validation.}
SENRI-Bench contains 10 self-recorded egocentric videos totaling 25.3 hours,
with 50 questions per video. We collected new recordings outside public video
repositories, which reduces the risk that they appeared in the pretraining
corpora of the evaluated models. Every video is captured in a single
uninterrupted take using a chest-mounted action camera.
The recordings span both indoor and outdoor environments, including shopping
malls, supermarkets, restaurants, museums, parks, campuses, streets, public
transit and so on.
The videos preserve physical transitions between places, for example, a walk from
a park to a subway station, movement among floors and stores in a mall, and
travel between indoor and outdoor regions. Then we
process the recordings
with EgoBlur~\citep{egoblur} to detect and blur visible faces and vehicle
license plates. The person who records a video manually writes its questions. This design gives
the annotator first-hand knowledge of the route, environment, and events,
supporting questions about spatial connections and intervening events. The
annotator constructs candidate
four-choice questions, assigns each to one of the six task types, and provides
one correct answer together with three challenging distractors. An independent verifier then checks each
question against the full video, ensuring that it is clearly worded, grounded
in visible evidence, and admits a single unambiguous answer. Questions with
ambiguous wording, overlapping answer choices, or insufficient visual evidence
are revised or removed to ensure the correctness and minimize the ambiguity.

\paragraph{Statistics and Visualization.}
SENRI-Bench comprises 500 four-choice questions over 10 videos, with 50
questions per video and 25.3 hours of footage in total (2.53 hours per video on
average). The benchmark contains 334 Event$\rightarrow$Space questions and 166
Space$\rightarrow$Event questions. The former comprise 89 localization, 90
relation reasoning, 86 object counting, and 69 navigation planning questions;
the latter comprise 88 RER and 78 trajectory event retrieval (TER) questions.
Correct-answer positions are balanced both within individual videos
and across the full benchmark, with 130, 130, 120, and 120 answers in positions
A--D, respectively. Figure~\ref{fig:task_distribution} visualizes this
hierarchical task distribution, the balanced correct-answer positions, and the
lexical composition of the questions and answer options.
Figure~\ref{fig:benchmark_examples} shows one representative example from each
task type and pairs every question with its visual evidence picked from the original long video in a temporal order.

\section{Benchmark Results}
\label{sec:benchmark_results}

\subsection{Experimental Setup}

\paragraph{Evaluated Methods.} We evaluate 28 model configurations spanning three paradigms for long-video
understanding. Direct perception covers both frontier proprietary
models and representative open-weight video models, including general-purpose
multimodal models~\citep{qwen3.5,qwen36_27b,qwen36_35b_a3b,qwen38,kimik3,gemini35flash,gemini37flash,gemini31pro,openai2025gpt5,openai2026gpt55} and models specialized for spatial or long-video
understanding~\citep{videochat3,internvideo3,cambrians}. Evidence
seeking covers systems that inspect query-relevant video regions on demand
through sequential or parallel tool use~\citep{videoo3,longvideor1,paravt,
videoarm}. Persistent memory covers systems that preprocess a video
into reusable text or visual memories before receiving a
question then perform multi-round retrieving and reasoning before answering~\citep{m3agent,lightomni,worldmm,egagent,magicvideo}.

\begin{table}[t]
    \centering
    \caption{
        Main results on SENRI-Bench (accuracy, \%). ES and SE denote
        Event$\rightarrow$Space and Space$\rightarrow$Event, respectively;
        L, NP, OC, RER, RR, and TER denote localization, navigation planning,
        object counting, region-conditioned event retrieval, relation reasoning,
        and trajectory event retrieval. $^{*}$ indicates that the
        inference backbone is further fine-tuned or post-trained by the
        corresponding method. Best and second-best results are shown in bold and underlined, respectively.
    }
    \label{tab:main_results}
    \scriptsize
    \setlength{\tabcolsep}{2.2pt}
    \renewcommand{\arraystretch}{0.99
    }
    \resizebox{\textwidth}{!}{%
    \begin{tabular}{@{}lllrrrrrrrrr@{}}
        \toprule
        \multirow{2}{*}{Approach} & \multirow{2}{*}{Method}
        & \multirow{2}{*}{Inference Model} & \multirow{2}{*}{Total}
        & \multicolumn{5}{c}{Event$\rightarrow$Space} & \multicolumn{3}{c}{Space$\rightarrow$Event} \\
        \cmidrule(lr){5-9} \cmidrule(lr){10-12}
        & & & & Avg. & L & NP & OC & RR & Avg. & RER & TER \\
        \midrule

        \multirow{15}{*}{\makecell[l]{Direct\\Perception}}
        & Gemini 3.1 Pro & Gemini 3.1 Pro
        & 47.4 & 43.1 & 56.2 & 37.7 & 37.2 & 40.0 & 56.0 & 55.7 & 56.4 \\
        & Gemini 3.5 Flash & Gemini 3.5 Flash
        & 49.8 & 45.5 & 56.2 & 40.6 & 32.6 & 51.1 & 58.4 & 60.2 & 56.4 \\
        & Gemini 3.7 Flash & Gemini 3.7 Flash
        & 52.2 & 48.5 & 61.8 & 50.7 & 32.6 & 48.9 & 59.6 & 58.0 & \textbf{61.5} \\
        & GPT-5 & GPT-5
        & 47.2 & 44.6 & 59.6 & 39.1 & 32.6 & 45.6 & 52.4 & 55.7 & 48.7 \\
        & GPT-5.5 & GPT-5.5
        & 48.2 & 46.7 & 60.7 & 44.9 & 31.4 & 48.9 & 51.2 & 50.0 & 52.6 \\
        & Kimi K3 & Kimi K3
        & 41.4 & 39.5 & 51.7 & 39.1 & 30.2 & 36.7 & 45.2 & 43.2 & 47.4 \\

        \cmidrule(lr){2-12}
        & Cambrian-S-7B & Qwen2.5-7B$^{*}$
        & 34.8 & 32.0 & 37.1 & 37.7 & 19.8 & 34.4 & 40.4 & 42.0 & 38.5 \\
        & Cambrian-S-7B-LFP & Qwen2.5-7B$^{*}$
        & 31.4 & 29.3 & 33.7 & 36.2 & 18.6 & 30.0 & 35.5 & 42.0 & 28.2 \\
        & InternVideo3-8B & Qwen3-VL-8B$^{*}$
        & 39.0 & 38.0 & 50.6 & 37.7 & 26.7 & 36.7 & 41.0 & 36.4 & 46.2 \\
        & VideoChat3-4B & Qwen3-4B$^{*}$
        & 40.8 & 40.4 & 44.9 & 46.4 & 32.6 & 38.9 & 41.6 & 39.8 & 43.6 \\
        & Qwen3.5-9B & Qwen3.5-9B
        & 41.6 & 41.6 & 53.9 & 39.1 & 31.4 & 41.1 & 41.6 & 44.3 & 38.5 \\
        & Qwen3.5-27B & Qwen3.5-27B
        & 43.8 & 41.6 & 48.3 & 43.5 & 32.6 & 42.2 & 48.2 & 44.3 & 52.6 \\
        & Qwen3.6-27B & Qwen3.6-27B
        & 43.8 & 41.0 & 51.7 & 44.9 & 29.1 & 38.9 & 49.4 & 43.2 & 56.4 \\
        & Qwen3.6-35B-A3B & Qwen3.6-35B-A3B
        & 45.6 & 46.1 & 51.7 & 43.5 & 34.9 & 53.3 & 44.6 & 44.3 & 44.9 \\
        & Qwen3.8-27B & Qwen3.8-27B
        & 41.0 & 39.5 & 48.3 & 37.7 & 37.2 & 34.4 & 44.0 & 45.5 & 42.3 \\

        \midrule
        \multirow{5}{*}{\makecell[l]{Evidence\\Seeking}}
        & LongVideo-R1 & Qwen3-8B$^{*}$
        & 39.0 & 36.5 & 39.3 & 27.5 & 33.7 & 43.3 & 44.0 & 40.9 & 47.4 \\
        & ParaVT-8B & Qwen3-VL-8B$^{*}$
        & 29.8 & 28.1 & 29.2 & 42.0 & 20.9 & 23.3 & 33.1 & 35.2 & 30.8 \\
        & Video-o3 & Qwen2.5-VL-7B$^{*}$
        & 40.6 & 40.1 & 48.3 & 43.5 & 29.1 & 40.0 & 41.6 & 37.5 & 46.2 \\
        & VideoARM & Qwen3.8-27B
        & 45.0 & 43.7 & 52.8 & 39.1 & 41.9 & 40.0 & 47.6 & 42.0 & 53.8 \\
        & VideoARM & Gemini 3.7 Flash
        & \textbf{59.0} & 56.0 & \underline{64.0} & \underline{56.5} & \underline{45.3} & 57.8 & \textbf{65.1} & \textbf{68.2} & \textbf{61.5} \\

        \midrule
        \multirow{8}{*}{\makecell[l]{Persistent\\Memory}}
        & M3-Agent & Qwen3-32B$^{*}$
        & 33.6 & 33.2 & 41.6 & 23.2 & 31.4 & 34.4 & 34.3 & 36.4 & 32.1 \\
        & Light-Omni & Qwen2.5-Omni-7B$^{*}$
        & 37.4 & 35.0 & 47.2 & 33.3 & 27.9 & 31.1 & 42.2 & 43.2 & 41.0 \\
        & EG-Agent & Qwen3.8-27B
        & 34.8 & 34.1 & 42.7 & 27.5 & 33.7 & 31.1 & 36.1 & 39.8 & 32.1 \\
        & EG-Agent & Gemini 3.7 Flash
        & 52.2 & 53.6 & \textbf{69.7} & 49.3 & \textbf{46.5} & 47.8 & 49.4 & 53.4 & 44.9 \\
        & WorldMM & Qwen3.8-27B
        & 47.6 & 42.8 & 46.1 & 31.9 & 36.0 & 54.4 & 57.2 & 61.4 & 52.6 \\
        & WorldMM & Gemini 3.7 Flash
        & \underline{58.8} & \underline{56.6} & 58.4 & \textbf{66.7} & 43.0 & \underline{60.0} & \underline{63.3} & \underline{67.0} & \underline{59.0} \\
        & MAGIC-Video & Qwen3.8-27B
        & 45.2 & 43.1 & 53.9 & 31.9 & 34.9 & 48.9 & 49.4 & 48.9 & 50.0 \\
        & MAGIC-Video & Gemini 3.7 Flash
        & 58.6 & \textbf{57.2} & 61.8 & 55.1 & 41.9 & \textbf{68.9} & 61.4 & \underline{67.0} & 55.1 \\

        \bottomrule
    \end{tabular}%
    }
\end{table}

\paragraph{Evaluation Protocol.}
Every configuration receives the same 500 questions and answer options. For
direct perception models, we uniformly sample 512 frames from each video and
provide them to the model together with the question. Cambrian-S-7B,
Cambrian-S-7B-LFP, InternVideo3-8B, and VideoChat3-4B are instead evaluated
using the default inference settings provided by their original repositories.
Most evidence seeking agents defaultly uses its original configurations, including its search
procedure, tool configuration, and
inference model. Most persistent memory systems use their native memory interface and retrieval-answer pipeline.
To maximize the fairness of comparison, we substitute the inference backbone of VideoARM, WorldMM, EG-Agent, and MAGIC-Video with Gemini 3.7 Flash and Qwen3.8-27B respectively while keeping the memories constructed by default settings.
Other methods use their own fine-tuned versions of models so that we keep their original inference backbone. Moreover, for methods that are or depend on models that can consume videos, we encodes the frames into videos to reduce token cost and avoids context-length limitations. We report the overall accuracy and the accuracy of each task type, and we also report the accuracy of each subtask (ES and SE).
The results are shown in Table~\ref{tab:main_results}.

\subsection{Main Findings}

\paragraph{Performance on SENRI.}
SENRI remains difficult for all three approaches. The best overall result is
59.0\%, and no baseline reaches 60\%. Direct perception models range from
31.4\% to 52.2\%, evidence seeking systems range from 29.8\% to 59.0\%, and
persistent memory systems range from 33.6\% to 58.8\%. Among direct perception
models, Gemini 3.7 Flash obtains 52.2\% and overall performance of proprietary models is better than open-source ones. VideoARM with Gemini reaches 59.0\%,
while WorldMM and MAGIC-Video with the same inference model reach 58.8\% and
58.6\%, suggesting that existing agentic approaches are not yet sufficient to fully address the challenges. 
These results also leave a difference between the two
directions of SENRI. ES accuracy is lower than SE accuracy in 25 of the 28
configurations. The strongest direct perception model scores 48.5\% on ES and
59.6\% on SE. VideoARM scores 56.0\% and 65.1\%, while WorldMM with Gemini
scores 56.6\% and 63.3\%. Only Qwen3.6-35B-A3B and EG-Agent with Gemini perform
better on ES, and Qwen3.5-9B obtains the same score in both directions. This suggests that under the current task composition, models struggle more with recovering spatial details from event cues than the opposite.

\begin{figure*}[t]
    \centering
    \includegraphics[width=\textwidth]{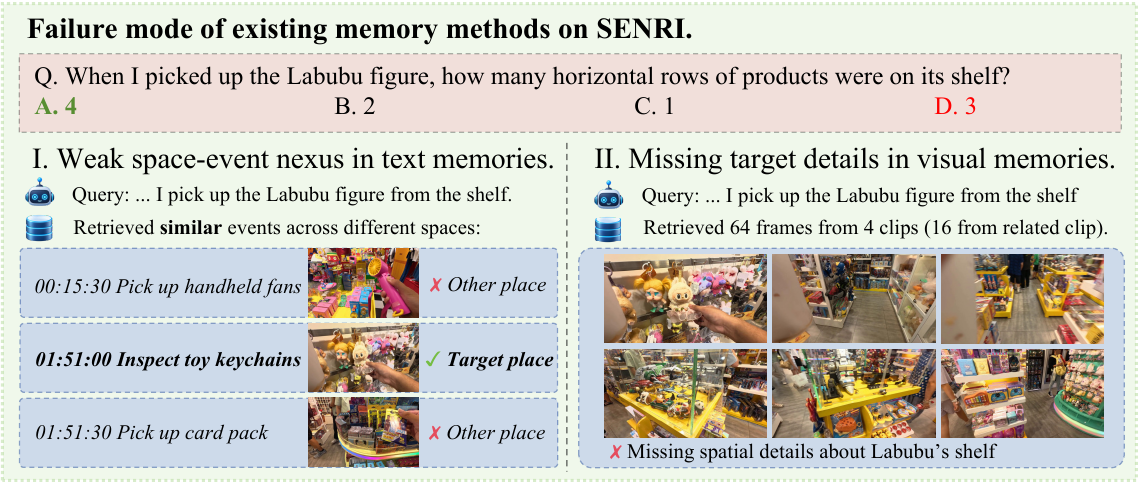}
    \vspace{-6mm}
    \caption{Representative failure modes of existing memory systems on
    SENRI: weak space-event nexus in text memories (left), and missing target details in visual memories (right).}
    \label{fig:memory_failure_modes}
\end{figure*}

\paragraph{Methods' performance differs across tasks.}
No method leads all six task types. Persistent memory methods achieve the best
results on the four ES tasks, but the leading method changes from one task to
another. EG-Agent with Gemini reaches 69.7\% on localization and 46.5\% on
object counting. MAGIC-Video with Gemini reaches 68.9\% on relation reasoning,
and WorldMM with Gemini reaches 66.7\% on navigation planning. Compared with
the best direct perception results, these scores improve localization by 7.9
percentage points, object counting by 9.3 points, relation reasoning by 15.6
points, and navigation planning by 16.0 points. Object counting nevertheless
remains the most difficult ES task, since even its best result is below 50\%.
The pattern differs for the two SE tasks. VideoARM with Gemini obtains the best
RER result at 68.2\%, which is 8.0 points above the best direct perception
model. On TER, VideoARM and direct Gemini 3.7 Flash tie at 61.5\%, followed by
WorldMM with Gemini at 59.0\%. The variation across tasks indicates that each
approach offers a different advantage and reveals the large room for improvement in SENRI-Bench.

\paragraph{Methods' performance depends on the inference model.}
Changing the inference model produces substantial gains even when the method
and its memories remain fixed. Replacing Qwen3.8-27B with Gemini 3.7
Flash raises VideoARM from 45.0\% to 59.0\%, WorldMM from 47.6\% to 58.8\%,
EG-Agent from 34.8\% to 52.2\%, and MAGIC-Video from 45.2\% to 58.6\%. The gains
range from 11.2 to 17.4 percentage points. They also appear in both directions. WorldMM improves by 13.8 points on ES and 6.1
points on SE. EG-Agent improves by 19.5 and 13.3 points, while MAGIC-Video
improves by 14.1 and 12.0 points. VideoARM gains 12.3 points on ES and 17.5
points on SE. These differences reflect that inference model plays a decisive role in long video understanding. This observation also motivates our later experiments,
which preserve the stored memories and retrieval interface while improving how
the retrieved evidence is verified and used.

\subsection{Failure Modes of Memory Systems}
\label{sec:memory_failure_analysis}

Former quantitative results show that existing memory systems still fall short of understanding space--event nexus 
in real-life videos. To better understand
limitations, we qualitatively analyze representative failure modes in Figure~\ref{fig:memory_failure_modes}. It illustrates two ways in which retrieved
memories can be relevant yet insufficient for answering an event-conditioned
spatial question. In this example, the event cue ``picking up a Labubu
figure'' must be linked to the corresponding shelf with 4 rows.

\paragraph{I. Weak space--event nexus in text memory.}
In existing memory methods, retrievers write queries to find relevant text memories but find it hard to model space--event relationships. For example, the system retrieves three superficially related merchandise interactions:
picking up handheld fans, inspecting toy keychains, and picking up a card pack.
However, the two picking-up events occur in other places, while the only result does not show the spatial details. The retrieved text therefore identifies plausible moments
without preserving the event--space nexus or the spatial attribute required by
the question. The ambiguity might cause misleading reasoning, or increases the numbers of reasoning rounds needed to further inspect the original video visually.

\paragraph{II. Missing target details in visual memory.}
Although existing methods can retrieve visual evidence, they often fail to capture the specific details needed to answer the question. As is shown in Figure~\ref{fig:memory_failure_modes}, these frames recover the surrounding store and even the
Labubu interaction, but none clearly shows the shelf configuration needed to
count its rows, since the sampling is too sparse to cover the necessary viewpoints that would reveal the shelf's structure. The system reaches the
correct temporal neighborhood, yet sparse visual evidence omits the decisive
details.

\section{SENRI-Agent}

\subsection{Method}
\label{sec:method}

\begin{figure}[t]
    \centering
    \includegraphics[width=\textwidth]{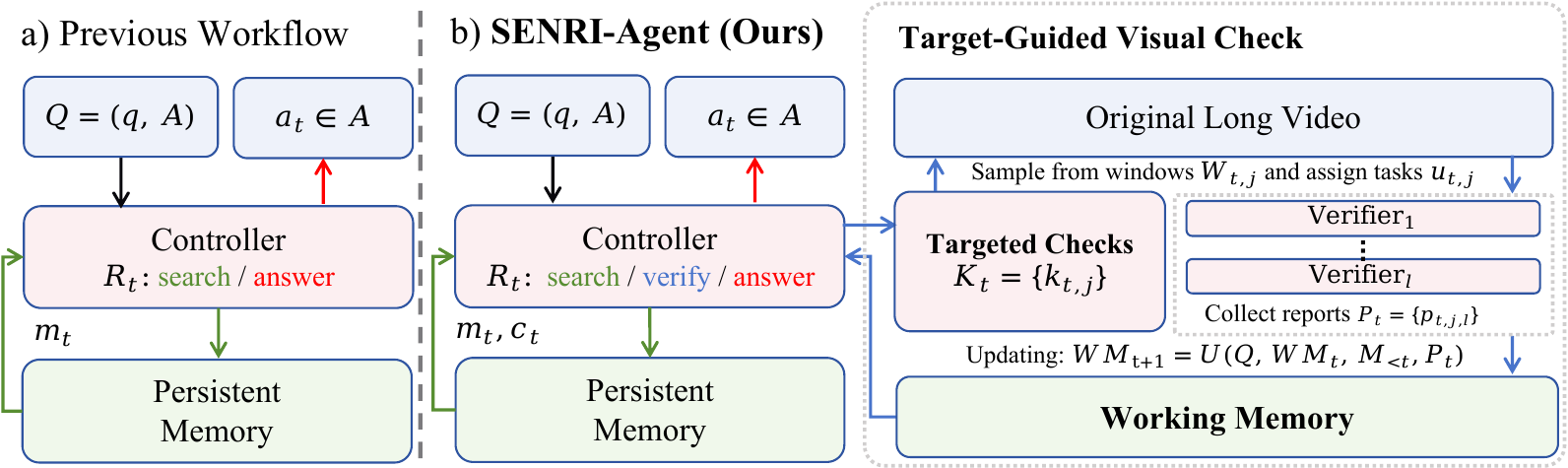}
    \vspace{-6mm}
    \caption{Overview of SENRI-Agent. (a) The previous workflow alternates
    between retrieving persistent memories and answering the question.
    (b) SENRI-Agent introduces a \texttt{verify} action that turns retrieved
    memories and visual candidates into target-guided checks over selected
    video windows. Parallel verifiers return reports with supporting evidence
    frames, which are integrated with memories from previous rounds into a
    bounded working memory for subsequent reasoning.}
    \label{fig:senri_agent}
\end{figure}

The failure modes mentioned above reveal the ambiguity of retrieved memories and motivate a new approach that goes beyond simple retrieval. Thus we propose \textbf{SENRI-Agent}, a controller workflow built upon existing constructed memories and augment the systems with a \textit{Target-Guided Visual Check (TGVC)} module. The controller uses the retrieved candidates to propose local visual check tasks for parallel verifiers and integrate the reports into a working memory to help with reasoning. The workflow is illustrated in Figure~\ref{fig:senri_agent}, which also shows the difference from the previous one.

\paragraph{Main Workflow.} Given a question and options $Q=(q,\mathcal{A})$, the controller chooses among three actions, \texttt{search}, \texttt{verify}, and \texttt{answer}, at each round $t\in\{1,2,\ldots,T\}$. If the controller selects \texttt{search}, the retriever returns $
\mathcal{M}_t=\{m_{t,i}\}_{i=1}^{n_t}, 
\,\,
\mathcal{C}_t=\{c_{t,i}\}_{i=1}^{n_t},
$
where $m_{t,i}$ is a textual memory and $c_{t,i}=([s_{t,i},e_{t,i}],d_{t,i})$ is a visual candidate registered from its time anchor which would be used by \texttt{verify}. Here $[s_{t,i},e_{t,i}]$ is its absolute interval in the source video and $d_{t,i}$ is its description. The accumulated sets are updated by $\mathcal{M}_{<t+1}=\mathcal{M}_{<t}\cup\mathcal{M}_t$ and $\mathcal{C}_{<t+1}=\mathcal{C}_{<t}\cup\mathcal{C}_t$. If the controller selects \texttt{answer}, it returns $a_t\in\mathcal{A}$ and terminates the workflow.
At the beginning of each round, its context is $\mathcal{H}_t
=
\left(
Q,
\mathcal{M}_{<t},
\mathcal{C}_{<t},
\mathrm{WM}_t
\right),$
where $\mathrm{WM}_t$ is the bounded multimodal working memory.
\paragraph{Target-Guided Visual Check.} If the controller chooses \texttt{verify}, it constructs $N_t$ visual checks:
\begin{equation}
\mathcal{K}_t
=
\{k_{t,j}\}_{j=1}^{N_t},
\qquad
k_{t,j}
=
(u_{t,j}, \mathcal{W}_{t,j}),
\end{equation}
where \(u_{t,j}\) is a structured verification task specifying the target occurrence and the visual fact to inspect, and \(\mathcal{W}_{t,j}\) contains the absolute-time windows selected from \(\mathcal{C}_{<t}\). The controller may expand or shorten the original candidate windows when constructing \(\mathcal{W}_{t,j}\).

For each check, the executor adaptively divides its verification windows into $L_{t,j}$ overlapping temporal cells, which are distributed to parallel verifiers for independent inspection. For cell $\ell$, the verifier produces a structured report and its supporting evidence frames:
\begin{equation}
p_{t,j,\ell}
=
(r_{t,j,\ell}, E_{t,j,\ell})
=
\mathcal{V}\left(
u_{t,j},
X_{t,j,\ell}
\right),
\end{equation}
where \(X_{t,j,\ell}\) denotes the timestamped visual observations from the cell, \(r_{t,j,\ell}\) summarizes the visually supported facts and any remaining ambiguity, and \(E_{t,j,\ell}\) contains the timestamped evidence frames supporting the report. This decomposition allows the model to inspect local details without uniformly adding a large number of frames to the controller context.

The verifier does not observe the answer options and reports only visually supported facts. The resulting verification outputs
$
\mathcal{P}_t
=
\left\{
p_{t,j,\ell}
\mid
1\leq j\leq N_t,
1\leq \ell\leq L_{t,j}
\right\}
$
are passed to the working-memory module, which retains the information needed for subsequent controller decisions.

\paragraph{Working Memory.}
Visual verification produces structured reports together with their supporting evidence frames. We maintain them as a bounded multimodal working memory
\begin{equation}
\mathrm{WM}_t
=
\left(
z_t,
\{p_i\}_{i\in\mathcal{S}_t}
\right)
=
\left(
z_t,
\{(r_i,E_i)\}_{i\in\mathcal{S}_t}
\right),
\end{equation}
where \(z_t\) is a concise summary, \(\mathcal{S}_t\) indexes the verification outputs currently retained in working memory, \(r_i\) is a selected verifier report, and \(E_i\) contains its supporting timestamped evidence frames.

After each \texttt{verify} action, the working memory is updated by the new verification reports \(\mathcal{P}_t\):
\begin{equation}
\mathrm{WM}_{t+1}
=
\mathcal{U}(Q,
\mathrm{WM}_t,
\mathcal{M}_{<t},
\mathcal{P}_t
),
\end{equation}
where \(\mathcal{M}_{<t}\) contains the memories exposed in all previous rounds, and \(\mathcal{P}_t\) contains the newly generated report--evidence pairs. The update refreshes the summary and selects a small subset of verification outputs that are most useful for subsequent decisions. Selected reports are preserved together with their original supporting evidence frames, while the summary records established facts, unresolved ambiguities, and evidence still needed. This bounded representation prevents the multimodal context from growing across rounds while retaining the evidence required for later reasoning.

  \subsection{Experiments}

  \textbf{Experimental Setup.} SENRI-Agent is built upon two representative persistent memory methods, WorldMM and MAGIC-Video by reusing their constructed memories and retrieval 
  interfaces. We evaluate SENRI-Agent on SENRI-Bench against the original workflows of the two
  baselines using two inference models, Qwen3.8-27B and Gemini 3.7 Flash. The
  maximum number of reasoning rounds is set to 5, followed by one answer round,
  as in the baselines. We keep 3 reports and 4 evidence frames per report in the working memory and set maximum 3 checks per verification.
  
  \textbf{Main Results.} Table~\ref{tab:senri_agent_results} shows that
  SENRI-Agent improves all four configurations using the same stored memories
  and retrieval interfaces. With Gemini 3.7 Flash, WorldMM improves from
  58.8\% to 62.6\% and MAGIC-Video from 58.6\% to 61.8\%. With Qwen3.8-27B,
  they improve from 47.6\% to 50.4\% and from 45.2\% to 47.8\%, respectively.
  The overall gains range from 2.6 to 3.8 percentage points, with improvements
  on both ES and SE across all four configurations. This supports the benefit
  of targeted verification and evidence retention across the tested systems.

  \begin{table}[t]
    \centering
    \caption{Results of SENRI-Agent on SENRI-Bench across different models. All are accuracy (\%).}
    \label{tab:senri_agent_results}
    \scriptsize
    \setlength{\tabcolsep}{3.0pt}
    \renewcommand{\arraystretch}{0.9}
    \resizebox{\textwidth}{!}{%
    \begin{tabular}{@{}lllrrrrrrrrr@{}}
        \toprule
        Memory system & Inference model & Workflow & Total
        & ES & L & NP & OC & RR & SE & RER & TER \\
        \midrule
        \multirow{2}{*}{WorldMM}
        & \multirow{2}{*}{Qwen3.8-27B}
        & Original & 47.6 & 42.8 & 46.1 & 31.9 & 36.0 & 54.4 & 57.2 & 61.4 & 52.6 \\
        & & SENRI-Agent & 50.4 & 46.1 & 56.2 & 34.8 & 38.4 & 52.2 & 59.0 & 63.6 & 53.8 \\
        \cmidrule(lr){2-12}
        \multirow{2}{*}{WorldMM}
        & \multirow{2}{*}{Gemini 3.7 Flash}
        & Original & 58.8 & 56.6 & 58.4 & 66.7 & 43.0 & 60.0 & 63.3 & 67.0 & 59.0 \\
        & & SENRI-Agent & 62.6 & 62.0 & 66.3 & 63.8 & 48.8 & 68.9 & 63.9 & 70.5 & 56.4 \\
        \cmidrule(lr){2-12}
        \multirow{2}{*}{MAGIC-Video}
        & \multirow{2}{*}{Qwen3.8-27B}
        & Original & 45.2 & 43.1 & 53.9 & 31.9 & 34.9 & 48.9 & 49.4 & 48.9 & 50.0 \\
        & & SENRI-Agent & 47.8 & 45.2 & 55.1 & 42.0 & 36.0 & 46.7 & 53.0 & 54.5 & 51.3 \\
        \cmidrule(lr){2-12}
        \multirow{2}{*}{MAGIC-Video}
        & \multirow{2}{*}{Gemini 3.7 Flash}
        & Original & 58.6 & 57.2 & 61.8 & 55.1 & 41.9 & 68.9 & 61.4 & 67.0 & 55.1 \\
        & & SENRI-Agent & 61.8 & 60.8 & 66.3 & 56.5 & 52.3 & 66.7 & 63.9 & 70.5 & 56.4 \\
        \bottomrule
    \end{tabular}%
    }
\end{table}

  \begin{table}[t]
    \centering
    \caption{
        Component ablations of SENRI-Agent on MAGIC-Video with Qwen3.8-27B. TG, PV, and WM denote target guidance, parallel verifiers, and working memory, respectively.}
    \label{tab:method_ablation}
    \scriptsize
    \setlength{\tabcolsep}{3.0pt}
    \renewcommand{\arraystretch}{0.95}
    \resizebox{\textwidth}{!}{%
    \begin{tabular}{@{}lccc|rrrrrrrrr@{}}
        \toprule
        Variant
        & TG & PV & WM
        & Total & ES & L & NP & OC & RR & SE & RER & TER \\
        \midrule
        SENRI-Agent
        & \ck & \ck & \ck
        & 47.8 & 45.2 & 55.1 & 42.0 & 36.0 & 46.7 & 53.0 & 54.5 & 51.3 \\
        w/o Target Guidance
        & \xk & \ck & \ck
        & 45.0 & 41.0 & 50.6 & 37.7 & 25.6 & 48.9 & 53.0 & 53.4 & 52.6 \\
        w/o Parallel Verifiers
        & \ck & \xk & \ck
        & 45.0 & 41.6 & 47.2 & 36.2 & 30.2 & 51.1 & 51.8 & 55.7 & 47.4 \\
        w/o Working Memory
        & \ck & \ck & \xk
        & 44.0 & 41.9 & 50.6 & 30.4 & 36.0 & 47.8 & 48.2 & 46.6 & 50.0 \\
        \bottomrule
    \end{tabular}%
    }
\end{table}

  \textbf{Ablation Studies.} Table~\ref{tab:method_ablation} reports component
  ablations on MAGIC-Video with Qwen3.8-27B. We remove target guidance by
  replacing structured verification tasks with a general observation prompt,
  remove parallel verifiers by assigning each check to a single verifier,
  and remove working memory by returning all accumulated reports and their
  supporting visual evidence to the controller. These changes reduce overall
  accuracy from 47.8\% to 45.0\%, 45.0\%, and 44.0\%, respectively,
  supporting the contribution of each component. Removing target guidance
  affects ES alone at the aggregate level, while removing parallel verifiers
  lowers accuracy in both directions. Removing working memory causes the
  largest overall drop, at 3.8 percentage points.

\section{Related Work}
\label{sec:related_work}

\paragraph{Spatial Understanding Benchmarks.}
Existing video benchmarks assess spatial intelligence from multiple perspectives. VSI-Bench~\citep{vsibench} evaluates spatial reasoning within individual
indoor scenes. VSI-Super~\citep{cambrians} concatenates recordings to extend
recall and counting. VSI-Super-Wild~\citep{vsisuperwild} uses continuous natural recordings, and
UCS-Bench~\citep{ucsbench} evaluates spatial reasoning under changes in user
position and orientation. LongSpace-Bench~\citep{longspace} evaluates spatial memory in room-tour videos. These tasks primarily characterize spatial configurations, motion, and object counts, whereas SENRI focuses on how such spatial information is grounded in and linked to events in self-recorded real-life videos.

\paragraph{Long Video Understanding Benchmarks.}
LVBench~\citep{lvbench}, LongVideoBench~\citep{longvideobench}, the Video-MME series~\citep{videomme,videommev2}, and MLVU~\citep{mlvu} benchmark long video understanding across diverse video sources and multiple tasks. HourVideo~\citep{hourvideo} extends such evaluation to hour-long videos and includes spatial reasoning. EgoSchema~\citep{egoschema} focuses on temporally extended reasoning over egocentric activities, while HippoVlog~\citep{hippomm} and EgoLifeQA~\citep{egolife} move toward understanding long-form daily-life experiences. More recently, EgoMemReason~\citep{egomemreason} and EgoMonth~\citep{egomonth} evaluate long-horizon memory over continuous experiences spanning days or weeks. SENRI centers its evaluation on the connection between space and events. Each video is a single uninterrupted take that preserves movement between spaces and events.

\paragraph{Agentic Long Video Understanding Methods.}
A growing body of work formulates long video understanding as an agentic process. One line of work performs evidence seeking, where an agent iteratively selects temporal regions, invokes multimodal tools, and refines its reasoning based on the acquired evidence~\citep{videoagent,videoo3,longvideor1,paravt,videoarm,videoseek,egor1}. Related agentic methods have also been explored for spatial intelligence, where spatial tools are used to accumulate geometric and semantic evidence across observations~\citep{sagent}. Another line builds persistent memory over videos, organizing visual experience into reusable episodic, semantic, multimodal, or structured representations that can support subsequent queries~\citep{m3agent,lightomni,worldmm,egagent,magicvideo,memdreamer,merit}. Our method builds on the established memories and uses the retrieved evidence for target-guided visual check in the original video.

\section{Conclusion}

In this work, we introduced SENRI, a new perspective on long video understanding that focuses on the nexus between events and their spatial contexts. We instantiated this problem in SENRI-Bench, a benchmark of long, uninterrupted, self-recorded egocentric videos with six task types spanning two complementary directions, Event$\rightarrow$Space and Space$\rightarrow$Event. Through an extensive evaluation of direct perception, evidence seeking, and persistent memory approaches, we found that existing systems still struggle on this task. Our analysis further revealed two representative limitations of current memory systems. Motivated by these observations, we proposed SENRI-Agent, which augments existing persistent memory systems with Target-Guided Visual Check and multimodal working memory, consistently improving their performance without modifying the underlying memories. We hope SENRI provides a useful testbed for studying how long video systems can move beyond existing abilities toward building memories that preserve the structure of lived experience.

\subsection*{AI use statement}

 In this work, we used generative AI tools to support qualitative and thematic data analysis, interpret results, provide feedback on research methodology or experiments, implement methods, and assist with translation. We have not used generative AI tools to generate
  synthetic datasets, develop theoretical models or conceptual frameworks, propose or refine hypotheses, or clean and reformat datasets, and tasks involving the formulation of mathematical claims and the development or writing of proofs are not applicable to this work.
  Additionally, we used generative AI tools to create or modify scientific figures, draft parts of the manuscript, summarize or analyze existing literature, explore research topics or identify gaps, brainstorm ideas, search for information, improve the readability of
  the manuscript, identify relevant literature, format references, and suggest the manuscript structure. We have reviewed all AI-assisted work. Specifically, we checked factual claims and references against their sources, reviewed interpretations against experimental
  results, and verified the correctness of AI-assisted code. We take responsibility for the final content of this work, including text, claims or artifacts produced with the aid of generative AI.

\subsection*{Ethics statement}

  SENRI-Bench contains egocentric recordings of everyday environments, which may capture bystanders and vehicle license plates. To reduce privacy risks, we use EgoBlur to detect and blur visible faces and license plates in the recordings. The benchmark focuses on the
  relationships between events and their spatial contexts and does not include tasks requiring the identification of individuals.

\subsection*{Reproducibility statement}

We describe the collection, annotation, and validation of SENRI-Bench in Section~\ref{sec:benchmark}. The experimental setup specifies the evaluated methods, inference models, and evaluation protocol. Section~\ref{sec:method} presents the SENRI-Agent workflow and its
  verification and working-memory components. The appendix provides additional evaluation settings, implementation details, and prompts, together with experiments on frame budgets and performance across individual videos.

\bibliography{iclr2027_conference}
\bibliographystyle{iclr2027_conference}

\appendix
\clearpage
\section{More Details of SENRI-Bench}

\subsection{More Questions, Options and Answers}

We provide more examples of questions, options and answers in SENRI-Bench.

Notably, some answer options indicate that a referenced entity or event is absent, encouraging models to verify the question’s premise against visual evidence. This design helps to test the model's ability to reason about the absence of certain elements in the video.

\begingroup
\small
\setlength{\tabcolsep}{3pt}
\renewcommand{\arraystretch}{1.15}

\begin{longtable}{@{}>{\raggedright\arraybackslash}p{0.12\textwidth}>{\raggedright\arraybackslash}p{0.27\textwidth}>{\raggedright\arraybackslash}p{0.51\textwidth}>{\centering\arraybackslash}p{0.04\textwidth}@{}}
\caption{Localization: ten selected SENRI-Bench questions. Answers are the original benchmark annotations.}\label{tab:examples_l}\\
\toprule
Example & Question & Options & Ans. \\
\midrule
\endfirsthead
\multicolumn{4}{l}{\textit{Localization (continued)}}\\
\toprule
Example & Question & Options & Ans. \\
\midrule
\endhead
\midrule
\multicolumn{4}{r}{\textit{Continued on next page}}\\
\endfoot
\bottomrule
\endlastfoot
\textbf{L-01} & Where was the trash can where I threw the ice cream cone box? & \textbf{A.} To the right of a pink-themed bus stop.\par \textbf{B.} A trash can inside the subway station.\par \textbf{C.} Opposite a green-themed bus stop.\par \textbf{D.} To the left of a blue-themed bus stop. & \textbf{A} \\[4pt]

\textbf{L-02} & Where did I pick up the guide booklet for this exhibition? & \textbf{A.} On the left side of the entrance to the first exhibition hall\par \textbf{B.} On the right side of the entrance to the first exhibition hall\par \textbf{C.} On the left side of the entrance to the second exhibition hall\par \textbf{D.} On the right side of the entrance to the second exhibition hall & \textbf{B} \\[4pt]

\textbf{L-03} & Where did I return the yellow power bank? & \textbf{A.} Inside the McDonald's by the river\par \textbf{B.} Inside the theater by the river\par \textbf{C.} At the subway entrance\par \textbf{D.} Inside the art museum by the river & \textbf{B} \\[4pt]

\textbf{L-04} & When I was storing my bag at the art museum, which locker did I store it in? & \textbf{A.} In the left cabinet, second from the left in the first row counting from bottom to top\par \textbf{B.} In the right cabinet, first from the left in the fifth row counting from bottom to top\par \textbf{C.} In the left cabinet, second from the left in the second row counting from bottom to top\par \textbf{D.} In the right cabinet, second from the left in the fourth row counting from bottom to top & \textbf{A} \\[4pt]

\textbf{L-05} & Where was I when I tasted the fruit flower tea? & \textbf{A.} I was at the fourth shop from the end under the bridge arch.\par \textbf{B.} I was at the first shop above the bridge arch.\par \textbf{C.} I was at the fourth shop from the end before entering the bridge arch.\par \textbf{D.} I was at the first shop under the bridge arch. & \textbf{C} \\[4pt]

\textbf{L-06} & Where did I hang my Messi doll? & \textbf{A.} On the platform by the window\par \textbf{B.} In the blue-themed store\par \textbf{C.} On a brown bench by the window\par \textbf{D.} On the bench next to the escalator & \textbf{C} \\[4pt]

\textbf{L-07} & When I was eating spaghetti, which floor was I on? & \textbf{A.} 3\par \textbf{B.} 1\par \textbf{C.} 2\par \textbf{D.} 4 & \textbf{C} \\[4pt]

\textbf{L-08} & When I parked my bicycle, where did I park it? & \textbf{A.} At the school gate, next to the subway station\par \textbf{B.} At the school gate, next to the bus station\par \textbf{C.} At the mall entrance, next to the subway station\par \textbf{D.} At the mall entrance, next to the bus station & \textbf{B} \\[4pt]

\textbf{L-09} & Where was I when I examined the race car? & \textbf{A.} B2\par \textbf{B.} 3rd floor\par \textbf{C.} 1st floor of the mall\par \textbf{D.} B1 & \textbf{D} \\[4pt]

\textbf{L-10} & When I took off my earphones, where was I? & \textbf{A.} Inside a pavilion\par \textbf{B.} On a rockery\par \textbf{C.} On a bridge in the middle of a pond, next to a pavilion\par \textbf{D.} None of the others are correct & \textbf{C} \\[4pt]

\end{longtable}

\begin{longtable}{@{}>{\raggedright\arraybackslash}p{0.12\textwidth}>{\raggedright\arraybackslash}p{0.27\textwidth}>{\raggedright\arraybackslash}p{0.51\textwidth}>{\centering\arraybackslash}p{0.04\textwidth}@{}}
\caption{Relation Reasoning: ten selected SENRI-Bench questions. Answers are the original benchmark annotations.}\label{tab:examples_rr}\\
\toprule
Example & Question & Options & Ans. \\
\midrule
\endfirsthead
\multicolumn{4}{l}{\textit{Relation Reasoning (continued)}}\\
\toprule
Example & Question & Options & Ans. \\
\midrule
\endhead
\midrule
\multicolumn{4}{r}{\textit{Continued on next page}}\\
\endfoot
\bottomrule
\endlastfoot
\textbf{RR-01} & When I was trying out pens, what was the spatial relationship between the shelf holding pens and the shelf holding correction tape? & \textbf{A.} The correction tape and the pens are on the same shelf, the former on the right and the latter on the left.\par \textbf{B.} The correction tape and the pens are on the same shelf, the former on the left and the latter on the right.\par \textbf{C.} The shelf for correction tape is closer to the open area, separated from the shelf for pens by a shelf of notebooks.\par \textbf{D.} The shelf for pens is closer to the open area, separated from the shelf for correction tape by a shelf of notebooks. & \textbf{D} \\[4pt]

\textbf{RR-02} & When I was viewing the exhibition, what was the spatial relationship between the video art installation and an abstract work related to a forest? & \textbf{A.} There was no video art installation at all\par \textbf{B.} Facing the video projection screen, the abstract work was behind me on the right\par \textbf{C.} Facing the video projection screen, the abstract work was directly behind me\par \textbf{D.} Facing the video projection screen, the abstract work was behind me on the left & \textbf{D} \\[4pt]

\textbf{RR-03} & When I bought the drink, what was the relative position of the checkout counter and the cabinet where I bought the drink? & \textbf{A.} The counter was on the same side as the cabinet.\par \textbf{B.} I did not buy a drink.\par \textbf{C.} I bought the drink from a vending machine.\par \textbf{D.} The counter was opposite the cabinet. & \textbf{D} \\[4pt]

\textbf{RR-04} & When I was picking up utensils, what were the relative positions of the fork and spoon to the cashier? & \textbf{A.} The fork was next to the cashier, and the spoon was next to the cashier\par \textbf{B.} The fork was next to the cashier, and the spoon was opposite the cashier\par \textbf{C.} The fork was opposite the cashier, and the spoon was next to the cashier\par \textbf{D.} The fork was opposite the cashier, and the spoon was opposite the cashier & \textbf{C} \\[4pt]

\textbf{RR-05} & When I bought the Messi doll, what was the relative position of the adjacent Messi and Ronaldo dolls? & \textbf{A.} Messi was to the right of Ronaldo\par \textbf{B.} There was no Messi\par \textbf{C.} Messi was to the left of Ronaldo\par \textbf{D.} There was no Ronaldo & \textbf{C} \\[4pt]

\textbf{RR-06} & When I was eating spaghetti, what was the relative position of the beverage counter to me? & \textbf{A.} To my rear left\par \textbf{B.} To my rear right\par \textbf{C.} To my front left\par \textbf{D.} To my front right & \textbf{C} \\[4pt]

\textbf{RR-07} & When I was buying braised food, what was the relative position between the shop I bought from and the yellow snack shop (from my perspective)? & \textbf{A.} The former is immediately next to the latter, on the latter's left\par \textbf{B.} The former is immediately next to the latter, on the latter's right\par \textbf{C.} The former is to the front right of the latter\par \textbf{D.} The former is to the front left of the latter & \textbf{A} \\[4pt]

\textbf{RR-08} & When I sat down to rest for the first time, was I closer to the nearest river or closer to the subway station? & \textbf{A.} Same distance\par \textbf{B.} The latter\par \textbf{C.} The former\par \textbf{D.} Cannot determine & \textbf{C} \\[4pt]

\textbf{RR-09} & When I ordered the milkshake, what was the relative position of the table where I finally sat down to me? & \textbf{A.} To my front left\par \textbf{B.} To my rear left\par \textbf{C.} To my rear right\par \textbf{D.} To my front right & \textbf{D} \\[4pt]

\textbf{RR-10} & When I was sitting and eating, what was the relative position of the cashier counter where I had just paid to me? & \textbf{A.} To my rear right\par \textbf{B.} To my rear left\par \textbf{C.} To my front right\par \textbf{D.} To my front left & \textbf{C} \\[4pt]

\end{longtable}

\begin{longtable}{@{}>{\raggedright\arraybackslash}p{0.12\textwidth}>{\raggedright\arraybackslash}p{0.27\textwidth}>{\raggedright\arraybackslash}p{0.51\textwidth}>{\centering\arraybackslash}p{0.04\textwidth}@{}}
\caption{Object Counting: ten selected SENRI-Bench questions. Answers are the original benchmark annotations.}\label{tab:examples_oc}\\
\toprule
Example & Question & Options & Ans. \\
\midrule
\endfirsthead
\multicolumn{4}{l}{\textit{Object Counting (continued)}}\\
\toprule
Example & Question & Options & Ans. \\
\midrule
\endhead
\midrule
\multicolumn{4}{r}{\textit{Continued on next page}}\\
\endfoot
\bottomrule
\endlastfoot
\textbf{OC-01} & When I was looking at the displayed books and other exhibits inside the Chinese-style building, how many glass display cabinets were in the room? & \textbf{A.} 4\par \textbf{B.} 3\par \textbf{C.} 2\par \textbf{D.} 8 & \textbf{D} \\[4pt]

\textbf{OC-02} & When I was trying out pens, how many pens were in the pen holder on my left (not on the shelf)? & \textbf{A.} 7\par \textbf{B.} 5\par \textbf{C.} 8\par \textbf{D.} 6 & \textbf{A} \\[4pt]

\textbf{OC-03} & When I was storing my bag at the art museum, how many compartments did the storage locker have? & \textbf{A.} 64\par \textbf{B.} 48\par \textbf{C.} 16\par \textbf{D.} 36 & \textbf{D} \\[4pt]

\textbf{OC-04} & When I observed the fish pond, how many green chairs were there in total around the fish pond? & \textbf{A.} 6\par \textbf{B.} 3\par \textbf{C.} 4\par \textbf{D.} 5 & \textbf{D} \\[4pt]

\textbf{OC-05} & After I took away the salad, how many boxes of salad were left on the shelf? & \textbf{A.} 6\par \textbf{B.} 3\par \textbf{C.} 4\par \textbf{D.} 5 & \textbf{C} \\[4pt]

\textbf{OC-06} & When I was eating spaghetti, how many plates were on my table? & \textbf{A.} 3\par \textbf{B.} 2\par \textbf{C.} 1\par \textbf{D.} 4 & \textbf{B} \\[4pt]

\textbf{OC-07} & When I picked up a tray of instant noodle cups, how many cups were there in total? & \textbf{A.} 2\par \textbf{B.} 3\par \textbf{C.} 1\par \textbf{D.} 4 & \textbf{B} \\[4pt]

\textbf{OC-08} & When I was smelling perfume, how many bottles of perfume were in that row in total? & \textbf{A.} 4\par \textbf{B.} 2\par \textbf{C.} 1\par \textbf{D.} 3 & \textbf{A} \\[4pt]

\textbf{OC-09} & When I was eating noodles, how many tables were there in the restaurant? & \textbf{A.} 9\par \textbf{B.} 5\par \textbf{C.} 11\par \textbf{D.} 7 & \textbf{C} \\[4pt]

\textbf{OC-10} & When I bought the sour plum drink (before purchasing), how many drinks were on the cart in total? & \textbf{A.} 15\par \textbf{B.} 13\par \textbf{C.} 19\par \textbf{D.} 17 & \textbf{D} \\[4pt]

\end{longtable}

\begin{longtable}{@{}>{\raggedright\arraybackslash}p{0.12\textwidth}>{\raggedright\arraybackslash}p{0.27\textwidth}>{\raggedright\arraybackslash}p{0.51\textwidth}>{\centering\arraybackslash}p{0.04\textwidth}@{}}
\caption{Navigation Planning: ten selected SENRI-Bench questions. Answers are the original benchmark annotations.}\label{tab:examples_np}\\
\toprule
Example & Question & Options & Ans. \\
\midrule
\endfirsthead
\multicolumn{4}{l}{\textit{Navigation Planning (continued)}}\\
\toprule
Example & Question & Options & Ans. \\
\midrule
\endhead
\midrule
\multicolumn{4}{r}{\textit{Continued on next page}}\\
\endfoot
\bottomrule
\endlastfoot
\textbf{NP-01} & From the Indian restaurant where I was last, to reach the conveyor-belt sushi restaurant, how should I walk? & \textbf{A.} Exit the restaurant, turn left, take the escalator down to the lowest floor, the sushi restaurant is on the right.\par \textbf{B.} Exit the restaurant, turn left, then make a left U-turn to go up the escalator, the sushi restaurant is on my right.\par \textbf{C.} Exit the restaurant, turn left, then make a left U-turn to go up the escalator, the sushi restaurant is on my left.\par \textbf{D.} Exit the restaurant, turn right, then make a right U-turn to go up the escalator, the sushi restaurant is on my right. & \textbf{B} \\[4pt]

\textbf{NP-02} & When I exited the art museum through the door near the river and wanted to take the subway, how did I get there? & \textbf{A.} Turn left from the door, turn left at the next corner, then ride a shared bike at the intersection, go straight all the way, it's on the left side\par \textbf{B.} Turn right from the door, walk straight along the river, then turn right at the intersection, go straight all the way, the subway entrance is on the left side\par \textbf{C.} Turn left from the door, then turn left at the art museum corner, then turn left again at the next corner, cross the crosswalk at the intersection, then go straight all the way, the subway entrance is on the left side\par \textbf{D.} Turn left from the door, then turn right at the next corner, cross the crosswalk at the intersection, then go straight all the way, turn right at a barber shop & \textbf{C} \\[4pt]

\textbf{NP-03} & From when I exited the underground mall to when I threw away my drink bottle, how did I walk? & \textbf{A.} Walked straight to near the bus stop, then turned right past many square-dancing aunties, then wandered counterclockwise in the park, and encountered many trash cans before completing half a circle.\par \textbf{B.} Walked straight to near the bus stop, then turned right past many square-dancing aunties, then wandered clockwise in the park past the fish pond, strolled by the pond, turned around and returned, then encountered trash cans.\par \textbf{C.} First turned around to the rear left, then turned right to near the bus stop, then turned right past many chairs, then wandered clockwise in the park past the fish pond, strolled by the pond, turned around and returned, then encountered trash cans.\par \textbf{D.} Walked straight to near the bus stop, then turned right past many square-dancing aunties, then wandered counterclockwise in the park past many trash cans, strolled by the pond, turned around and returned, then encountered trash cans. & \textbf{C} \\[4pt]

\textbf{NP-04} & When I was about to get on the subway, how did I get from the residential complex to the subway? & \textbf{A.} Turn left from the residential complex gate, then turn left at a parking lot gate, keep walking straight until reaching the roadside, then turn left\par \textbf{B.} Turn right from the residential complex gate, then turn left at a parking lot gate, keep walking straight until reaching the roadside, then turn left\par \textbf{C.} Walk from the residential complex gate to Starbucks, make a U-turn to the right at the Starbucks entrance, keep walking straight until reaching the roadside, then turn left.\par \textbf{D.} Turn left from the Starbucks entrance, keep walking straight until reaching the roadside, then turn left & \textbf{A} \\[4pt]

\textbf{NP-05} & When I got off the bicycle, if I wanted to go directly to the subway station without passing through the shopping mall, how should I go? & \textbf{A.} Make a U-turn along the direction I was last cycling, turn left under the elevated bridge, then keep going straight ahead\par \textbf{B.} Keep going straight ahead along the direction I was last cycling, turn left under the elevated bridge, then keep going straight ahead\par \textbf{C.} Make a U-turn along the direction I was last cycling, turn right under the elevated bridge, then keep going straight ahead\par \textbf{D.} Keep going straight ahead along the direction I was last cycling, turn right under the elevated bridge, then keep going straight ahead & \textbf{D} \\[4pt]

\textbf{NP-06} & How did I ride from buying braised food to encountering the school gate for the first time? & \textbf{A.} Immediately turn left, then turn right, then turn left, then turn right, then turn right\par \textbf{B.} Immediately turn left, then turn left, then turn left, then turn right, then turn left\par \textbf{C.} Immediately turn left, then turn right, then turn left, then turn right, then turn left\par \textbf{D.} Immediately turn left, then turn left, then turn left, then turn right, then turn right & \textbf{A} \\[4pt]

\textbf{NP-07} & After I got off the bus, how did I walk to the park entrance? & \textbf{A.} Turn left onto the sidewalk, then turn left to cross the road, then turn left to go up onto the football field platform; after going up, go down from the right side to enter the park area\par \textbf{B.} Turn left onto the sidewalk, then turn left to cross the road, then turn right to go up onto the football field platform; after going up, go down from the right side to enter the park area\par \textbf{C.} Turn left onto the sidewalk, then turn left to cross the road, then turn left to go up onto the football field platform; after going up, go down from the left side to enter the park area\par \textbf{D.} Turn left onto the sidewalk, then turn right to cross the road, then turn right to go up onto the football field platform; after going up, go down from the right side to enter the park area & \textbf{B} \\[4pt]

\textbf{NP-08} & How did I walk during my transfer? & \textbf{A.} Turned left after exiting the car door, made a right U-turn to go upstairs, turned left to go upstairs again, then walked forward and to the left\par \textbf{B.} Turned left after exiting the car door, made a right U-turn to go upstairs, turned left to go upstairs again, then walked forward and to the right\par \textbf{C.} Turned right after exiting the car door, made a right U-turn to go upstairs, turned left to go upstairs again, then walked forward and to the right\par \textbf{D.} Turned left after exiting the car door, made a right U-turn to go upstairs, turned right to go upstairs again, then walked forward and to the right & \textbf{B} \\[4pt]

\textbf{NP-09} & When I came out of the toy store, if I wanted to walk to the subway entrance where I finally took the subway, how should I walk to get there faster? & \textbf{A.} Turn left after exiting and just keep walking along the road.\par \textbf{B.} Turn left after exiting, walk until you reach the bank, then turn right.\par \textbf{C.} Turn right after exiting and just keep walking along the road.\par \textbf{D.} Turn left after exiting, walk until you reach the bank, then turn left. & \textbf{C} \\[4pt]

\textbf{NP-10} & When I took the subway back, how did I get from the subway entrance to the platform entrance for the line I wanted to take? & \textbf{A.} First went down the escalator, passed the queuing iron railings, then swiped the ticket gate, then walked straight ahead, then went up the escalator, then turned left, turned right at the green billboard, and finally turned left to go down the escalator\par \textbf{B.} First went down the escalator, passed the queuing iron railings, then swiped the ticket gate, then walked straight ahead, then went up the escalator, then turned left, turned right at the yellow billboard, and finally turned left to go down the escalator\par \textbf{C.} First went down the escalator, passed the queuing iron railings, then swiped the ticket gate, then walked straight ahead, then went up the escalator, then turned left, and finally turned left to go down the escalator\par \textbf{D.} First went down the escalator, passed the queuing iron railings, then swiped the ticket gate, then walked straight ahead, then turned left, turned right at the yellow billboard, and finally turned left to go down the escalator & \textbf{B} \\[4pt]

\end{longtable}

\begin{longtable}{@{}>{\raggedright\arraybackslash}p{0.12\textwidth}>{\raggedright\arraybackslash}p{0.27\textwidth}>{\raggedright\arraybackslash}p{0.51\textwidth}>{\centering\arraybackslash}p{0.04\textwidth}@{}}
\caption{Region-Conditioned Event Retrieval: ten selected SENRI-Bench questions. Answers are the original benchmark annotations.}\label{tab:examples_rer}\\
\toprule
Example & Question & Options & Ans. \\
\midrule
\endfirsthead
\multicolumn{4}{l}{\textit{Region-Conditioned Event Retrieval (continued)}}\\
\toprule
Example & Question & Options & Ans. \\
\midrule
\endhead
\midrule
\multicolumn{4}{r}{\textit{Continued on next page}}\\
\endfoot
\bottomrule
\endlastfoot
\textbf{RER-01} & What did I do when I was at the entrance of a Mexican restaurant? & \textbf{A.} I took a number\par \textbf{B.} The waiter led me into the restaurant\par \textbf{C.} I asked the waiter\par \textbf{D.} I flipped through the menu & \textbf{D} \\[4pt]

\textbf{RER-02} & What did I do on the second-floor platform of the art museum that connects to the outside? & \textbf{A.} I drank a beverage\par \textbf{B.} I used binoculars to look at the opposite side\par \textbf{C.} I walked two circles around it\par \textbf{D.} I flipped through the art exhibition brochure & \textbf{D} \\[4pt]

\textbf{RER-03} & What did I do next to the children's slide? & \textbf{A.} I turned a turntable.\par \textbf{B.} I bought a bottled drink.\par \textbf{C.} I shoveled sand with a shovel.\par \textbf{D.} I touched the slide. & \textbf{A} \\[4pt]

\textbf{RER-04} & What did I do between the two orange lifebuoys? & \textbf{A.} Ate an ice cream cone\par \textbf{B.} Drank cola\par \textbf{C.} Threw away the bottle\par \textbf{D.} Drank plain water & \textbf{D} \\[4pt]

\textbf{RER-05} & What did I do by the fish pond? & \textbf{A.} Watched others feed the fish\par \textbf{B.} Fed the fish\par \textbf{C.} Threw away trash\par \textbf{D.} Sat and rested & \textbf{A} \\[4pt]

\textbf{RER-06} & What did I do when I was at the furniture store? & \textbf{A.} I bought a charging cable\par \textbf{B.} I tried to charge my phone\par \textbf{C.} I did nothing\par \textbf{D.} I played with the action camera & \textbf{B} \\[4pt]

\textbf{RER-07} & What did I do between the two stone lions? & \textbf{A.} I accepted a leaflet\par \textbf{B.} I put on my earphones\par \textbf{C.} I turned on my electric fan\par \textbf{D.} None of the others are correct & \textbf{A} \\[4pt]

\textbf{RER-08} & When I was next to the beverage machine, what was the last drink I poured? & \textbf{A.} Lemon tea\par \textbf{B.} Sprite\par \textbf{C.} Cola\par \textbf{D.} Fanta & \textbf{D} \\[4pt]

\textbf{RER-09} & What did I do next to the checkout counter in the supermarket? & \textbf{A.} I asked about the price\par \textbf{B.} None of the above\par \textbf{C.} I handed over paper money\par \textbf{D.} I brought up the payment QR code & \textbf{D} \\[4pt]

\textbf{RER-10} & What did I do on the second floor of the dining area? & \textbf{A.} I bought matcha\par \textbf{B.} I looked at what food was available and then turned around and went downstairs\par \textbf{C.} I sat down to rest\par \textbf{D.} None of the above & \textbf{B} \\[4pt]

\end{longtable}

\begin{longtable}{@{}>{\raggedright\arraybackslash}p{0.12\textwidth}>{\raggedright\arraybackslash}p{0.27\textwidth}>{\raggedright\arraybackslash}p{0.51\textwidth}>{\centering\arraybackslash}p{0.04\textwidth}@{}}
\caption{Trajectory Event Retrieval: ten selected SENRI-Bench questions. Answers are the original benchmark annotations.}\label{tab:examples_ter}\\
\toprule
Example & Question & Options & Ans. \\
\midrule
\endfirsthead
\multicolumn{4}{l}{\textit{Trajectory Event Retrieval (continued)}}\\
\toprule
Example & Question & Options & Ans. \\
\midrule
\endhead
\midrule
\multicolumn{4}{r}{\textit{Continued on next page}}\\
\endfoot
\bottomrule
\endlastfoot
\textbf{TER-01} & What did I encounter when I went from the restroom entrance to the Chinese-style building with the exhibition? & \textbf{A.} I passed by an ice cream stand\par \textbf{B.} I climbed over a rockery\par \textbf{C.} I encountered many children feeding the goldfish in the pond\par \textbf{D.} Several performers in ancient costumes passed in front of me & \textbf{D} \\[4pt]

\textbf{TER-02} & What happened on the way from the Pokémon card vending machine to the MUJI store? & \textbf{A.} I bought Pokémon cards\par \textbf{B.} I browsed a LEGO store\par \textbf{C.} I caught a doll\par \textbf{D.} I saw a claw machine but didn't play & \textbf{D} \\[4pt]

\textbf{TER-03} & On the way from McDonald's to the art museum, what did I do? & \textbf{A.} I returned a power bank\par \textbf{B.} I stopped to watch a street singer\par \textbf{C.} I threw a plastic bottle into a trash can\par \textbf{D.} I bought a sports drink & \textbf{A} \\[4pt]

\textbf{TER-04} & Between leaving the temple and crossing the street for the first time, what did I see? & \textbf{A.} A man was filming the pagoda\par \textbf{B.} A man was filming a green building\par \textbf{C.} A woman was taking a selfie\par \textbf{D.} A young person was flying a drone & \textbf{B} \\[4pt]

\textbf{TER-05} & After walking to the end along the canal side and then turning around to walk back along the opposite side, what did I encounter? & \textbf{A.} I stopped and observed a reptile experience center.\par \textbf{B.} Someone pitched flower tea to me.\par \textbf{C.} I drank water 5 times.\par \textbf{D.} I did a toy gun target shooting activity. & \textbf{A} \\[4pt]

\textbf{TER-06} & What did I see while walking from the ice cream vending machine to the bridge? & \textbf{A.} A sanitation worker riding a bicycle\par \textbf{B.} Someone driving a red vehicle\par \textbf{C.} A sanitation worker riding a tricycle\par \textbf{D.} Someone driving a white vehicle & \textbf{C} \\[4pt]

\textbf{TER-07} & What happened from the time I got on the shared bike until I arrived at the dormitory? & \textbf{A.} Someone asked me how to get to the subway station\par \textbf{B.} The toothbrush almost fell off the bike\par \textbf{C.} The potato chips almost fell off the bike\par \textbf{D.} My phone fell on the ground & \textbf{C} \\[4pt]

\textbf{TER-08} & What happened from the time I got on the bicycle until I turned left and then went straight? & \textbf{A.} I was overtaken from the right by a person wearing an orange hat and riding an e-bike\par \textbf{B.} I was overtaken from the right by a person wearing a yellow hat and riding an e-bike\par \textbf{C.} I was overtaken from the left by a person wearing a yellow hat and riding an e-bike\par \textbf{D.} I was overtaken from the left by a person wearing an orange hat and riding an e-bike & \textbf{C} \\[4pt]

\textbf{TER-09} & What happened while I went down the ladder on the right side of the large football field, turned left to pass the small football field, walked along the edge of the park, then turned left to enter the park and kept walking straight in? & \textbf{A.} I took a photo for three elderly people\par \textbf{B.} I scrutinized the red flowers by the roadside\par \textbf{C.} I bought water from a vending machine\par \textbf{D.} None of the above is correct & \textbf{B} \\[4pt]

\textbf{TER-10} & When I passed by a blue mobile pop-up shop, then turned left and then right into a roofed alley, what did I see while walking forward? & \textbf{A.} A green paper bag fell on the ground\par \textbf{B.} Someone tried to sell me gold products\par \textbf{C.} There was a person riding on a rolling suitcase.\par \textbf{D.} None of the above & \textbf{C} \\[4pt]

\end{longtable}

\endgroup

  \subsection{More Evaluation Details}
  \label{app:more_evaluation_details}

  We provide additional details on model inputs, inference
  settings,
  and scoring. Our evaluation covers three ways of
  accessing long-video
  evidence: direct video perception, on-demand evidence
  seeking, and
  persistent memory. Each configuration is evaluated on the
  same
  500 underlying questions, with four answer options per
  question.

  \paragraph{Direct video perception.}
  For general purpose direct perception models, the main
  evaluation
  uses a budget of 512 uniformly sampled frames per video.
  The sampled
  observations are supplied together with the question and
  answer
  options, allowing the model to reason directly over
  visual input.
  For evaluations using preconstructed video carriers, we
  verify that
  each carrier contains exactly 512 sampled frames. The
  open weight
  Qwen baselines use non-thinking greedy decoding, with the
  native
  video processing interface configured for at most 512
  frames.

  We retain the released inference settings for the
  specialist
  direct-perception systems, including Cambrian-S-7B,
  Cambrian-S-7B-LFP, InternVideo3-8B, and VideoChat3-4B.
  Their visual sampling and processing procedures therefore
  follow
  their respective implementations. These results evaluate
  the
  available systems under their intended inference
  procedures.

  \paragraph{Evidence-seeking systems.}
  Evidence-seeking methods access video observations
  through their
  released search or inspection interfaces. The inference backbone reported in
  Table~\ref{tab:main_results} is the model that makes the
  final
  answer decision; a system may additionally use other
  models
  within its tools. For example, LongVideo-R1 uses a Qwen3-
  8B
  controller together with Qwen2.5-VL-72B and Qwen2.5-VL-
  32B
  video tools.

  For VideoARM, we keep other configurations
  fixed
  when comparing Qwen3.8-27B and Gemini 3.7 Flash.
  This controls the search setup across inference
  backbones,
  although their generated decisions and resulting
  execution
  trajectories may still differ.

  \paragraph{Persistent-memory systems.}
  Persistent-memory methods construct reusable memories
  before
  receiving the benchmark questions. At evaluation time,
  the
  question-answering workflow accesses these memories
  through the
  method's retrieval interface. For the paired Qwen3.8-27B
  and
  Gemini 3.7 Flash evaluations of WorldMM, EG-Agent, and
  MAGIC-Video,
  we reuse the same method specific memories and retrieval
  implementations. Memory is not rebuilt separately for
  each
  inference backbone. Consequently, these paired
  comparisons
  measure how different inference models use a fixed memory
  resource within each method.

  Memory construction models are distinguished from the
  inference
  backbones listed in the results table. For example, M3-
  Agent
  additionally uses Qwen2.5-Omni-7B during memory
  construction.
  Such components remain part of the evaluated system even
  when
  they are not the model responsible for selecting the
  final answer.

  \paragraph{Metrics and aggregation.}
  Each category accuracy is computed using the number of
  questions
  in that category as its denominator. Overall accuracy is
  computed
  directly over the complete benchmark. We retain
  integer
  correct counts during aggregation and round percentages
  to
  one decimal place only after division. This ensures that
  the
  reported overall and subset results are derived
  consistently
  from the final question-level predictions. Reported differences in
  percentage points are calculated from the displayed accuracies.

\section{More Details of SENRI-Agent}
\label{app:method_details}

\subsection{Structured Target-Guided Tasks}
\label{app:target_task}

Following the notation in Section~\ref{sec:method}, a visual check is
$k_{t,j}=(u_{t,j},\mathcal{W}_{t,j})$. We instantiate its structured task as
\begin{equation}
    u_{t,j}=\bigl(g_{t,j},o_{t,j},\mathcal{R}_{t,j}\bigr),
    \qquad
    \mathcal{R}_{t,j}=(\rho_{t,j,1},\ldots,\rho_{t,j,K_{t,j}}),
\end{equation}
where $g_{t,j}$ is the \emph{target-occurrence context}, $o_{t,j}$ is one
\emph{local observation objective}, and $\mathcal{R}_{t,j}$ is an ordered
list of \emph{observable requirements}. These components correspond
to the fields \texttt{target\_context}, \texttt{objective}, and
\texttt{requirements}, respectively. 

The verifier receives $u_{t,j}$ and timestamped visual observations
$X_{t,j,\ell}$ without the original question or answer options. In its
requirement list, the target context is assigned identifier \texttt{R1};
the conditions in $\mathcal{R}_{t,j}$ follow as \texttt{R2}, \texttt{R3},
and so on. For each identifier, the verifier is required to return its finding and
  citation. Conditions without
a nonempty finding and valid supporting citations are returned as evidence
gaps.

Each window in $\mathcal{W}_{t,j}$ specifies a candidate reference and absolute
start and end times in its source video. Candidates are for references and the controller may choose other intervals within the
valid source-video range when supported by accumulated evidence.

\subsection{Adaptive Temporal Cells}
\label{app:temporal_cells}

For each check, the controller also specifies a sampling rate \(f_{t,j}\)
and a frame budget \(B_{t,j}\). The supported ranges are
\(0.05\leq f_{t,j}\leq4\) frames per second and
\(8\leq B_{t,j}\leq64\) frames. If the frame budget is omitted, the sparse
and dense sampling presets use \(8\) and \(16\) frames, respectively. These
sampling parameters govern execution and are separate from the semantic
task \(u_{t,j}\).

For clarity, fix \((t,j)\) and write its \(H\) windows as

$$
\mathcal{W}_{t,j}=\{[a_h,b_h]\}_{h=1}^{H},
$$

ordered by start time, with \(D_h=b_h-a_h>0\). We suppress candidate
references in this notation. For each window, define the nominal cell span
and the corresponding number of temporal cells as
\begin{equation}
\tau_h=\min\!\left(D_h,\frac{B_{t,j}}{f_{t,j}}\right),
\qquad
L_h=
\max\!\left(1,\left\lceil\frac{2D_h}{\tau_h}-1\right\rceil\right).
\label{eq:temporal_cell_count}
\end{equation}
Here, \(\tau_h\) is the longest nominal interval that can be sampled at rate
\(f_{t,j}\) without exceeding the frame budget \(B_{t,j}\). The cell count
\(L_h\) is chosen so that adjacent cells overlap by half their length while
covering the full window. A sufficiently short window therefore requires
only one cell.

Given \(L_h\), let
\begin{equation}
\delta_h=\frac{2D_h}{L_h+1},
\qquad
I_{h,r}=\left[
a_h+(r-1)\frac{\delta_h}{2}\,;
a_h+(r-1)\frac{\delta_h}{2}+\delta_h
\right],
\quad r=1,\ldots,L_h.
\label{eq:temporal_cell_intervals}
\end{equation}
The first cell starts at \(a_h\), the last ends at \(b_h\), and adjacent
cells have \(50\%\) overlap when \(L_h>1\). For \(L_h=1\), the cell is the
original window. Windows are subdivided independently, so separate requested
intervals are never joined across unrequested temporal gaps.

The total number of verifier tasks for the check is
\begin{equation}
L_{t,j}=\sum_{h=1}^{H}L_h.
\end{equation}
Cells from all windows are then ordered chronologically and indexed by
\(\ell=1,\ldots,L_{t,j}\).

Every cell inherits the same semantic task \(u_{t,j}\), sampling rate
\(f_{t,j}\), and frame cap \(B_{t,j}\), and yields one report--evidence pair
$
p_{t,j,\ell}=(r_{t,j,\ell},E_{t,j,\ell}).
$
In this temporal-cell execution mode, the frame cap applies independently
to each cell.
Consequently, a check may inspect up to \(L_{t,j}B_{t,j}\) sampled frames,
including repeated observations in overlapping regions.

\subsection{Key Prompt Excerpts}
\label{app:method_prompts}

The following passages reproduce key instructions from the controller, verifier, and working memory updater. We omit runtime context
and portions of the output schemas for readability. 

\paragraph{Controller.}
The controller receives $Q$, accumulated retrieved memories and candidates,
and the current working memory $\mathrm{WM}_t$. Its system prompt includes:
\begin{quote}\small
You are the planning controller for a long-video multiple-choice system.
Use accumulated persistent-memory and visual evidence to choose one allowed
action. For search, provide one standalone memory query.
\end{quote}
The verification instructions specify:
\begin{quote}\small
For verify, assign 1-3 independent small visual tasks with a target context,
one objective, visible requirements, and adaptive whole-video absolute
windows. The verifier receives only that task, timestamps, and frames.
Candidate refs identify the source; use only their listed
\texttt{valid\_absolute\_sec} range. For answer, provide one choice.
\end{quote}
The prompt also directs subsequent decisions using verifier feedback:
\begin{quote}\small
Make each check a small choice-blind job. Use the verifier's natural-language
ambiguity and requested next evidence when choosing a later action.
Keep the rationale short.
\end{quote}
The controller returns a JSON action containing the search query, visual
checks, or selected answer as appropriate. Each check contains the three
fields defining $u_{t,j}$, its windows, and its sampling parameters.

\paragraph{Visual verifier.}
For native-video input, the verifier system prompt includes:
\begin{quote}\small
You are a choice-blind visual observation worker. Complete only the supplied
local task from the short videos; you do not receive the original question
or choices. For each requirement, report the visible finding and cite moments
on the supplied video's local timeline. Explain ambiguity in plain language
and say what visible evidence is still needed.
\end{quote}
The user message supplies the local observation objective, target-occurrence
context, numbered visible conditions, and the video for the cell. The
required JSON fields include \texttt{observation}, \texttt{explanation},
\texttt{ambiguity\_reason}, \texttt{next\_evidence\_needed},
\texttt{evidence\_moments}, \texttt{requirement\_findings}, and
\texttt{requirement\_evidence\_moments}.
Sampled frames are presented chronologically in a video carrier at one
frame per second; this presentation rate is distinct from the source sampling
rate $f_{t,j}$. The citation instruction is:
\begin{quote}\small
Cite \texttt{offset\_sec} on this supplied carrier timeline; do not convert
it to an absolute source timestamp.
\end{quote}
The executor maps valid citations back to the sampled source timestamps to
construct the supporting evidence frames $E_{t,j,\ell}$.

\paragraph{Working-memory updater.}
The updater receives the question and options as fixed prompt context,
along with $\mathrm{WM}_t$, $\mathcal{M}_{<t}$, and $\mathcal{P}_t$.
Its system instructions include:
\begin{quote}\small
Maintain working memory by faithfully summarizing and selecting visual
reports for a long-video planner. The question and choices identify which
distinctions matter and how to organize the note; verifier reports remain
authoritative for what they observed. Write a concise note (aim for at most
250 words) focused on the evidence relevant to the question: what the reports
establish, what remains unresolved, and what additional evidence would
resolve it.
\end{quote}
The evidence-preservation constraints include:
\begin{quote}\small
Faithfully preserve all material evidence, qualifications, uncertainty,
viewing scope, and provenance from the reports. Do not infer, reinterpret,
recount, reconcile, or resolve conflicting evidence, and never state a
likely answer or preferred choice. When reports disagree, attribute each
claim to its report ID instead of producing a new combined judgment.
\end{quote}
Additional instructions separate retrieval claims from observations:
\begin{quote}\small
Images may help select reports and preserve provenance, but must not revise
report findings. Keep retrieval claims separate from visually established
facts, and keep missing target-occurrence evidence explicit. Select existing
report IDs whose original observations and images are most useful to retain;
you may select none. Do not invent report IDs.
\end{quote}
The available selection pool consists of previously retained reports and
the latest verification batch. The updater returns JSON with
\texttt{note} and \texttt{selected\_report\_ids}, defining $z_{t+1}$ and
$\mathcal{S}_{t+1}$, respectively. The selected reports and their associated
evidence frames are copied unchanged into $\mathrm{WM}_{t+1}$.

\section{More Experiments}

\subsection{Effect of Number of Frames Sampled}
\label{app:effect_num_frames}

\begin{figure}[t]
    \centering
    \includegraphics[width=\linewidth]{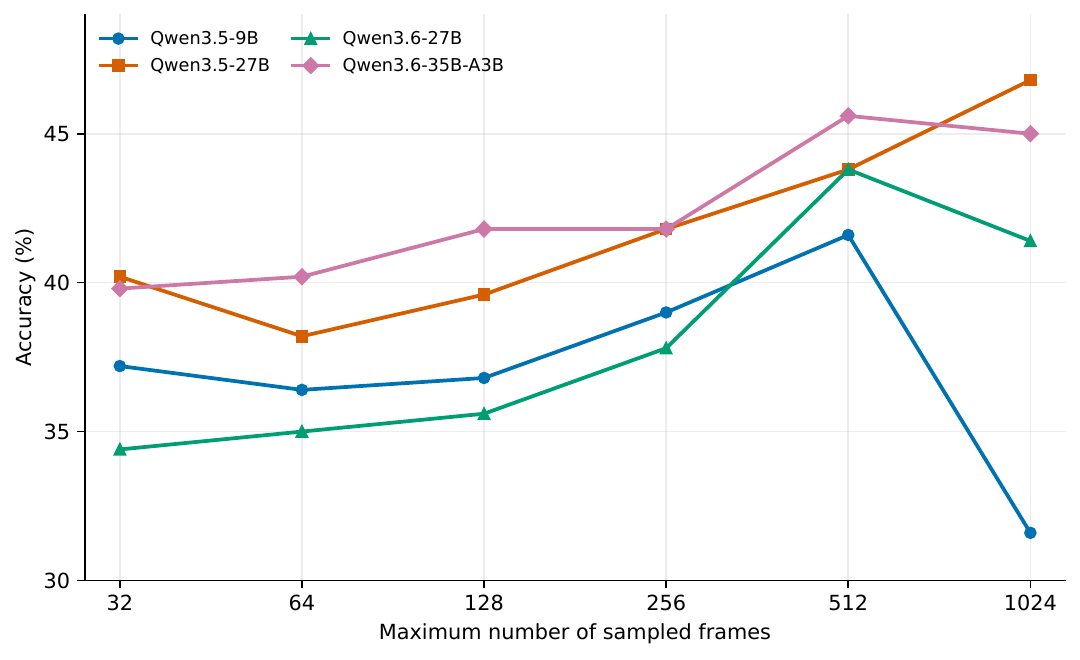}
    \caption{Accuracy on SENRI-Bench as the maximum number of sampled
    frames increases for four Qwen models.}
    \label{fig:effect_num_frames}
\end{figure}

We evaluate four Qwen models on all 500 SENRI-Bench questions with maximum
frame budgets of 32, 64, 128, 256, 512, and 1,024, using greedy decoding
with thinking disabled. Figure~\ref{fig:effect_num_frames} shows that all
four models obtain higher accuracy at 512 frames than at 32 frames, with
gains of 3.6 to 9.4 percentage points. The improvement varies across
models: Qwen3.6-27B rises from 34.4\% to 43.8\%, while Qwen3.5-27B rises
from 40.2\% to 43.8\%. Several models also lose accuracy at intermediate
budgets, so the benefit of additional frames is not uniform.

Increasing the budget from 512 to 1,024 frames raises Qwen3.5-27B from
43.8\% to 46.8\%, but lowers accuracy for the other three models. The comparison
therefore shows that additional frames can improve accuracy, but the largest budget
does not consistently give the best result and justifies the 512-frame budget used in our main experiments.

\subsection{Per-Video Analysis}
\label{app:per_video_analysis}

We examine how the gains in Table~\ref{tab:senri_agent_results} are
distributed across videos. For each memory system and inference model,
we compare the accuracy of the original workflow and SENRI-Agent on each
video. Each comparison covers the same 500 questions, with 50 questions
per video.

\begin{figure}[t]
    \centering
    \includegraphics[width=\linewidth]{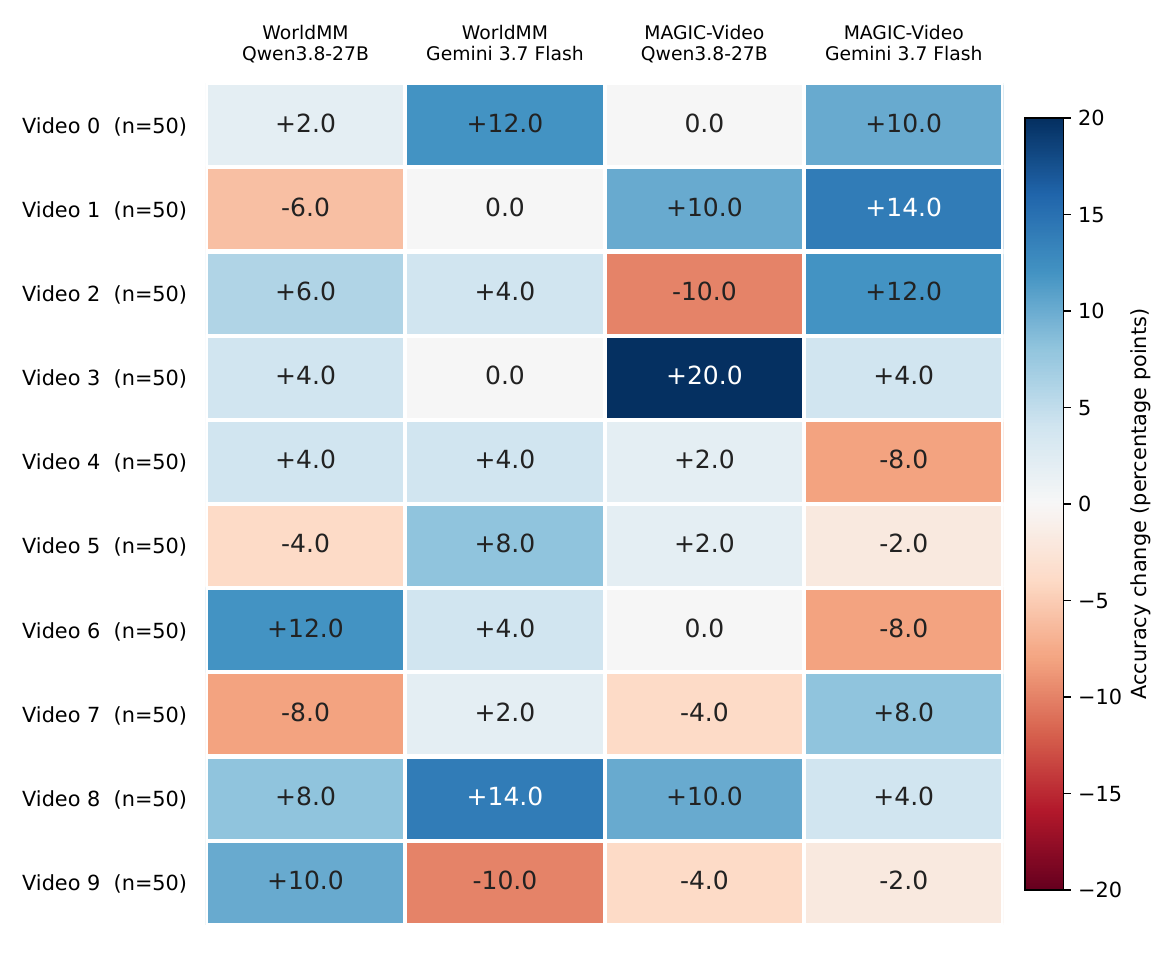}
    \caption{Accuracy change of SENRI-Agent relative to the original
    workflow for each video, in percentage points. Each cell compares
    predictions on the same 50 questions. Blue indicates an improvement,
    red a decrease, and white no change. Videos are indexed from 0 to 9.}
    \label{fig:per_video_gain}
\end{figure}

Figure~\ref{fig:per_video_gain} shows that SENRI-Agent improves WorldMM
on seven of the ten videos with either inference model. With MAGIC-Video,
it improves five videos, ties two, and declines on three with Qwen3.8-27B;
with Gemini 3.7 Flash, it improves six and declines on four. The magnitude
of the gains also varies. For MAGIC-Video with Qwen3.8-27B, Videos 1 and 3
gain 10 and 20 percentage points, respectively, while Video 2 loses
10 points. Thus, the overall gains extend to multiple videos, but some
recordings contribute much more than others.

\subsection{Comparison with No Visual Input}
\label{app:no_visual_input}

We evaluate Qwen3.8-27B using only the question and its four answer options,
without video input or retrieved memories. The model answers 173 of the
500 questions correctly (34.6\%), compared with the 25\% expected accuracy
of uniform random guessing. Table~\ref{tab:no_visual_results} reports the
results alongside configurations that use video evidence.

With the same inference model, direct perception reaches 41.0\%, exceeding
the no-visual-input baseline by 6.4 percentage points. SENRI-Agent reaches
47.8\% on MAGIC-Video and 50.4\% on WorldMM, exceeding this baseline by
13.2 and 15.8 points, respectively, with higher accuracy on all six task
types. These comparisons show the advantage of the evaluated workflows
that use video evidence over answering from the question and options alone.
The additional gains of SENRI-Agent over the original memory workflows are
2.6 points for MAGIC-Video and 2.8 points for WorldMM
(Table~\ref{tab:senri_agent_results}).

\begin{table}[htbp]
    \centering
    \caption{Comparison with a no-visual-input baseline on SENRI-Bench
    (accuracy, \%). All evaluated settings use Qwen3.8-27B as the inference
    model. The no-visual-input baseline receives only the question and answer
    options. Uniform random guessing gives an expected accuracy of 25\%.}
    \label{tab:no_visual_results}
    \scriptsize
    \setlength{\tabcolsep}{3pt}
    \renewcommand{\arraystretch}{1.08}
    \resizebox{\textwidth}{!}{%
    \begin{tabular}{@{}lrrrrrrrrr@{}}
        \toprule
        Setting & Total & ES & L & NP & OC & RR & SE & RER & TER \\
        \midrule
        Random guessing (expected)
        & 25.0 & 25.0 & 25.0 & 25.0 & 25.0 & 25.0 & 25.0 & 25.0 & 25.0 \\
        Question and options only
        & 34.6 & 35.0 & 36.0 & 33.3 & 33.7 & 36.7 & 33.7 & 34.1 & 33.3 \\
        \midrule
        Direct perception (512 frames)
        & 41.0 & 39.5 & 48.3 & 37.7 & 37.2 & 34.4 & 44.0 & 45.5 & 42.3 \\
        SENRI-Agent on MAGIC-Video
        & 47.8 & 45.2 & 55.1 & 42.0 & 36.0 & 46.7 & 53.0 & 54.5 & 51.3 \\
        SENRI-Agent on WorldMM
        & 50.4 & 46.1 & 56.2 & 34.8 & 38.4 & 52.2 & 59.0 & 63.6 & 53.8 \\
        \bottomrule
    \end{tabular}%
    }
\end{table}

\section{Limitations and Future Work}

\paragraph{Benchmark scope.}
SENRI-Bench contains 500 manually annotated questions over 25.3 hours of
continuous egocentric video. The ten recordings cover several everyday
environments, but broader coverage of locations and activities would help
assess generalization. The four-choice format supports consistent scoring
across methods; it does not evaluate whether a model can independently
describe an event and its spatial context in a free-form answer. We chose 4-choice design 
by following the common practice in existing work on long video understanding benchmarks.
Expanding
the recordings and adding questions that require free-form answers are natural
extensions of the current benchmark and are good directions for future research.

\paragraph{Comparing ES and SE.}
The two directions contain different task types: ES covers localization,
relation reasoning, object counting, and navigation planning, while SE
covers two forms of event retrieval. The lower ES accuracy in 25 of the
28 configurations therefore describes performance under this task composition;
it does not establish that Event$\rightarrow$Space is intrinsically harder.
For example, object counting is a challenging category included only in ES.
Future evaluations could pair questions in both directions around the same
events and locations, with comparable evidence requirements and question
difficulty, to better separate the effects of direction and task composition.

\paragraph{Method evaluation.}
We did not cover all possible configurations of evaluated methods. And we did not cover all possible methods available. But we chose representative ones. More models' performance on SENRI-Bench remains interesting future work.
We evaluate SENRI-Agent with two persistent memory systems and two inference
models. Its improvements across these four configurations support the use of
targeted verification, while its applicability to other memory architectures
remains to be tested. And its performance on other datasets are also valuable future directions.
However, the current benchmark and method are designed to study the space--event nexus in long egocentric videos, and we leave the exploration of other video types to future work.

\end{document}